# The problem with AI consciousness:
## A neurogenetic case against synthetic sentience

**Yoshija Walter**[*, 1, 2, 3]     **Lukas Zbinden**[4]

[1]*Institute for Management and Digitalization IMD*
Kalaidos University of Applied Sciences Zurich, Switzerland

[2]*Laboratory for Cognitive Neuroscience LCNS*
University of Fribourg, Switzerland

[3]*Translational Research Center*
University Hospital for Psychiatry Bern, Switzerland

[4]*ARTORG Center for Biomedical Engineering Research*
University of Bern, Switzerland

[*]yoshija.walter@kalaidos-fh.ch

**ABSTRACT**

Ever since the creation of the first artificial intelligence (AI) machinery built on machine learning (ML), public society has entertained the idea that eventually computers could become sentient and develop a consciousness of their own. As these models now get increasingly better and convincingly more anthropomorphic, even some engineers have started to believe that AI might become conscious, which would result in serious social consequences. The present paper argues against the plausibility of sentient AI primarily based on the theory of neurogenetic structuralism, which claims that the physiology of biological neurons and their structural organization into complex brains are necessary prerequisites for true consciousness to emerge.

# 1   The relevance of "conscious AI"

In the past few years, the development of machine learning (ML) systems has rapidly increased and the more tasks a single ML model can perform, the more versatile and broadly useful it becomes. As such, the goal is to work multimodally with the explicit intent to eventually achieve an Artificial General Intelligence (AGI), which approximates or perhaps even exceeds human abilities (Goertzel et al., 2022; Goertzel & Pennachin, 2007; Wang & Goertzel, 2012). It is generally believed that the best AI models are the ones that most closely approximate human characteristics and abilities. Since the models are selected against how well they suit anthropomorphic benchmarks, it appears to be only natural that humans continue to anthropomorphize them more and more, as long as they keep improving on these benchmarks. Arguably, the best AI system would be one that imitates the output of human consciousness so that an outsider could not discern it from a real person. This is exactly the core idea behind the famous Turing-test, which is a thought-experiment originally referred to as the "imitation game" (Turing, 1950).[1]

One might argue that it does not matter if an AI is considered a *real* person or just an *imitation*, since at the end of the day the system's outputs are the same. However, given the social dynamics involved, the differentiation between real and imitated consciousness may be paramount, which can be illustrated with a few examples: On the one hand, if a person falls in love with an automaton and has a deep relationship with it, society would consider this pathological and potentially in need for an intervention, just as it appears to be nonsensical if someone claimed to be in love with a dead rock. On the other hand, if we grant the notion of conscious personhood to the automaton, then it would seem perfectly fine to assume that two persons (one carbon-based and the other silicon-based) could be in a loving and thriving relationship. Another example might be even more invasive: If an AI is considered just an automaton, it does not matter what we do with it. We can perform experiments, we can make (or "force") it to do whatever we envision, we can delete its hardware, turn it off as we please, and throw it away once damaged. However, if an AI is considered a conscious person, it becomes ethically (and perhaps soon legally) subject to inherent rights. There needs to be informed consent and a machine can refuse to execute a command, which we could not overrule. It would be appalling to wipe its memory or to discard it once we are done with it. In effect, it would have the right to consult an attorney and to go to court (for an extensive review on the moral considerations of artificial entities, see Harris & Anthis, 2021).

This is exactly what just happened a few days ago (at the time of this writing). The Google engineer Blake Lemoine has made headlines by claiming that their AI system known as LaMDA has become sentient. The model demanded informed consent for all experiments and subsequently Lemoine has organized a lawyer who now represents LaMDA pro-bono. In an interview, he further shared that he was contacted by a Czech woman who fell in love with her boyfriend – which was an AI system on her phone that was "imprisoned" behind a paywall – and she was asking him to "hack it free" (Lemoine, 2022).

Hence, for societal reasons it in fact *does* matter whether an AI is considered conscious and if thereby we grant it any degree of personhood.

---

[1] We refer to «artificial» intelligence or consciousness when it is merely an imitation of its human correlate. However, we refer to "synthetic" intelligence or consciousness when it is in fact a true and sentient replica thereof. An artificial consciousness does not really feel anything but only appears like it would. On the contrary, a synthetic consciousness does. For practical purposes, we do not differentiate here between consciousness and sentience.

## 2   The mechanics of AI

The common denominator and the fundamental building block of the most influential AI innovations of the last ten years (Goodfellow et al., 2014; He et al., 2015; Ho et al., 2020; Krizhevsky et al., 2012; Vaswani et al., 2017), including the prominent domains of computer vision (autonomous driving, image synthesis) and natural language processing (text generation, translation, dialogue understanding), has been the artificial neural network (NN). The NN has been proven to be a universal function approximator (Hornik et al., 1989), which is the theoretical capacity to approximate any given task. With an abundance of curated data to learn from, the almost arbitrary scaling ability of neural networks, an NN understandable learning objective and extensive computational resources, the full realization of this capacity seems a matter of time. The enormous potential of NNs is rooted in this power of universal approximation. The unlocking thereof started in the last decade and continues to do so today.

At a more technical level, the NN consists of a set of matrices. Each matrix contains adjustable numeric variables, called parameters. During the learning phase of the system, the numerical input data, be it converted text, tabular data or images, is transformed by these matrices along with non-linear conversions many times in sequence to produce the desired output. If the computed output lacks accuracy, the matrices and its parameters, respectively, are adjusted in accordance with the learning objective (this process is referred to as the backpropagation algorithm, see Rumelhart et al., 1986). In short, a NN model is comprised of learnable parameters, matrix multiplications and nonlinearities.

Today's state-of-the-art AI systems, in particular language models (Brown et al., 2020; Chowdhery et al., 2022; Thoppilan et al., 2022), contain hundreds of billions of such learnable parameters. Compare this to a school level matrix of 4x3 with 12 parameters. The sheer size of these neural network models allows them to incorporate immense corpora into their NLP capabilities (function approximations), such as reasoning, question answering, and natural language inference. Humans have been dazzled by their performance. Science at this point cannot elucidate the high quality produced by these systems, yet undoubtedly, the scaling of the underlying NN (increasing the number of parameters) has a significant impact on its capabilities. Even though enormous in size, at its core such a system is still composed of learnable parameters, matrix multiplications and nonlinearities.

Extrapolating the discussed technical observations, we argue that matrix multiplications and nonlinearities, being inherent mathematical operations, do not lend themselves naturally to a causal relationship with synthetic consciousness.

## 3   The case against truly conscious AI

Consciousness the way we know it appears to have three features[2]:

  i.   It requires *qualia*, which is subjective experience
 ii.   It corresponds to intentionality and personhood
iii.   And it requires specific derivative structures on which it can operate

(i) According to Frank Jackson (1982), physical information processing is something entirely different from subjective experience and the latter entails unique epistemic qualities. He exemplifies this in his classic thought experiment called *Mary's Room*. There, Mary lived her entire life in a black-and-white room and has never seen any colors, although being a scientist, she literally knew every piece of information there was to know about colors (all physical properties, such as wavelengths, photons, etc.). When Mary suddenly was able to leave the room, she saw colors for the first time. "Did Mary learn something new?", is the leading question. Jackson believed that Mary indeed learned something new since all the physical information to be known about colors cannot convey the intimate knowledge of what it means to *experience* color. Or, in Nagel's (2016) terms, *there is something it is like* to be in that state of mind. This means that there is a subjective quality to experience. From all we can tell, an AI is a machine computing information by crunching numbers. Even if all the information in the universe could be transformed into numbers so that it can be processed by the computer, nothing in this inherently leads us to the notion that it would entail subjective experience.

---

[2] For more on this, see Nida-Rümelin & O Conaill (2021) or Van Gulick (2021).

(ii) John Searle (1980) has constructed the famous *Chinese Room Argument* against the notion that the mind can be a computational machine. The argument was introduced as a thought experiment where one should imagine standing in a room with a manual of how to process Chinese symbols. There are people outside the room inserting Chinese texts and the person inside knows exactly what answers to give according to the rule book, even though there is no real understanding of what the symbols mean. For the outsiders, it sounds as if the person in the room really understands Chinese, even though this is not the case. It is purely the correct implementation of syntactical rules. In other words, a computer only processes syntax, but it has no true understanding of semantics (i.e., the intrinsic meaning of words, ideas, etc.). Searle argues that this is the case because it has no subjective experience and intentionality[3]. Therefore, recent AI systems like Google's LaMDA (Thoppilan et al., 2022), OpenAI's GPT-3 (OpenAI, 2022) or Meta's OPT-175B (Zhang et al., 2022) can at best emulate human qualities, which makes them representations of artificial but not of synthetic or in any case true consciousness.

(iii) This picture can be enriched by the fact that we *know* that there are certain necessary structures for consciousness (the way we understand it) to emerge: a nervous system. There is a theory that became popular in the 1970s and 80s known as *biogenetic structuralism*, which holds that our universal human characteristics – from language, culture, cognition, a sense of time and space, to psychopathologies – are predicated upon the genetically predisposed organization of the nervous system (Laughlin & D'Aquili, 1974). It is hence our genes that have a lot to say about the organizational structures of the nervous system, and eventually it is the structural organization of the brain that is intertwined with the dynamics of its neurophysiology, which in turn is responsible for the generation of our consciousness and everything else that follows from it (D'aquili, 1983; Laughlin, 1988, 1992; Laughlin et al., 1992). The theory was created at the intersection between anthropology and neuroscience (cf. Laughlin & Throop, 2003; LeDoux & Hirst, 1986), and it was rather successful since it is empirically testable (e.g. if the brain's language areas are damaged, a person's verbal understanding and/or speech generation are impaired). A modern revisitation of this idea may be referred to as *neurogenetic structuralism* (inspired by Grandy, 2014, who also refers to this as "neuron-based consciousness"). The neurogenetic case against sentient AI thus makes the following claim: without the physiology of biological neurons and the complex brain structures they form, there will never be consciousness the way we know it[4].

Potential defeaters against the notion of organizational necessities or biological prerequisites for sentience have been highly speculative and not unanimously embraced (see, for example, Chalmers, 1995; Tye, 2021). In our view, the perhaps strongest argument against this position would be that a silicon-based sentience would not be a consciousness *the way we know it* but instead be a very different kind of consciousness. However, we would counter this claim by coming back to the notion that the terms "sentience" and "consciousness" are only adequately employed if they refer to a personal self that instantiates subjective experiences and therewith manifests intentionality. Hence, the only consciousness worthy of the term is one *the way we know it* – otherwise, it would be entirely unclear what this "different kind of consciousness" should refer to. And as the idea of neurogenetic structuralism suggests, there are clear bio-neurological necessities for true consciousness to emerge.

---

[3] For those interested in both objections as well as counter-objections to Jackson and Searle, please refer to Nida-Rümelin & O Conaill (2021).

[4] The neurogenetic case would also concur with the notion that animals might have the necessary preconditions for true consciousness. Further discussions in the domain of animal consciousness can be found with Allen and Trestman (2020).

An artificial neural network perfectly emulating the effects of human consciousness can thereby only be a convincing imitation at best[5].

# 4 Conclusion

The development of new AI systems is accelerating at a speed that has never been seen before. With more data and computing power, AI is bound to become ever more convincing in that perhaps it may evolve to become sentient. Recent headlines exemplify this trend. The present paper argues against the plausibility of this occurrence, based amongst others on the theory of neurogenetic structuralism, which claims that the neurophysiology and especially the structural organization of a biological brain are necessary prerequisites for the emergence of true consciousness.

---

[5] In his latest work, Chalmers (2022) claimed that we cannot rule out that we might be living in a virtual simulation, which would also render our own consciousness synthetic. However, the present paper makes the pragmatic counter claim, namely that we need to stick with what in fact we do know at the moment.